%% file: main.tex
\documentclass[10pt,twocolumn,letterpaper]{article}

\usepackage{3dv}
\usepackage{times}
\usepackage{epsfig}
\usepackage{graphicx}
\usepackage{amsmath}
\usepackage{amssymb}

\usepackage{multirow}
\usepackage[inline]{enumitem}
\usepackage{dsfont}
\usepackage[table]{xcolor}
\usepackage{multirow}
\usepackage{pifont}
\usepackage{algorithm}
\usepackage{algpseudocode}
\usepackage{romannum}
\usepackage{caption}
\usepackage{subcaption}
\usepackage{fix2col}
\usepackage{diagbox}
\usepackage{xspace}
\usepackage{tabularx,booktabs}
\usepackage{blindtext}
\usepackage{rongyu}

\usepackage[pagebackref=true,breaklinks=true,colorlinks,bookmarks=false]{hyperref}

\threedvfinalcopy 

\def\threedvPaperID{51} 

\ifthreedvfinal\fi
\begin{document}

\title{Monocular 3D Reconstruction of Interacting Hands \\ via Collision-Aware Factorized Refinements}

\author{
	Yu Rong$^{1}$ \hspace{9pt} Jingbo Wang$^{1}$ \hspace{9pt} Ziwei Liu$^{2}$ \hspace{9pt} Chen Change Loy$^{2}$\\
	{$^{1}$The Chinese University of Hong Kong} \hspace{10pt}
	{$^{2}$S-Lab, Nanyang Technological University}\\
	{\tt\small \{ry017, wj020\}@ie.cuhk.edu.hk \{ziwei.liu, ccloy\}@ntu.edu.sg}
}

\maketitle
\thispagestyle{empty}

\input{sections/abstract.tex}
\input{sections/introduction.tex}

\input{sections/related_work.tex}
\input{sections/method.tex}

\input{sections/experiment.tex}

\input{sections/conclusion.tex}
\clearpage
\input{sections/supp.tex}

{\small
\bibliographystyle{ieee_fullname}
\bibliography{long,mocap}
}

\end{document}

%% file: sections/abstract.tex

\begin{abstract}
\vspace{-0.26cm}

3D interacting hand reconstruction is essential to facilitate human-machine interaction and human behaviors understanding.
Previous works in this field either rely on auxiliary inputs such as depth images or they can only handle a single hand if monocular single RGB images are used. 
Single-hand methods tend to generate collided hand meshes, when applied to closely interacting hands, since they cannot model the interactions between two hands explicitly.
In this paper, we make the first attempt to reconstruct 3D interacting hands from monocular single RGB images.
Our method can generate 3D hand meshes with both precise 3D poses and minimal collisions.
This is made possible via a two-stage framework. Specifically, the first stage adopts a convolutional neural network to generate coarse predictions that tolerate collisions but encourage pose-accurate hand meshes.
The second stage progressively ameliorates the collisions through a series of factorized refinements while retaining the preciseness of 3D poses.
We carefully investigate potential implementations for the factorized refinement, considering the trade-off between efficiency and accuracy.
Extensive quantitative and qualitative results on large-scale datasets such as InterHand2.6M demonstrate the effectiveness of the proposed approach.
Code and demo are available at~\url{https://penincillin.github.io/ihmr_3dv2021}.

\vspace{-0.35cm}
\end{abstract}

%% file: sections/introduction.tex

\section{Introduction}

\begin{figure}[t]
	\begin{center}
		\includegraphics[width=\linewidth]{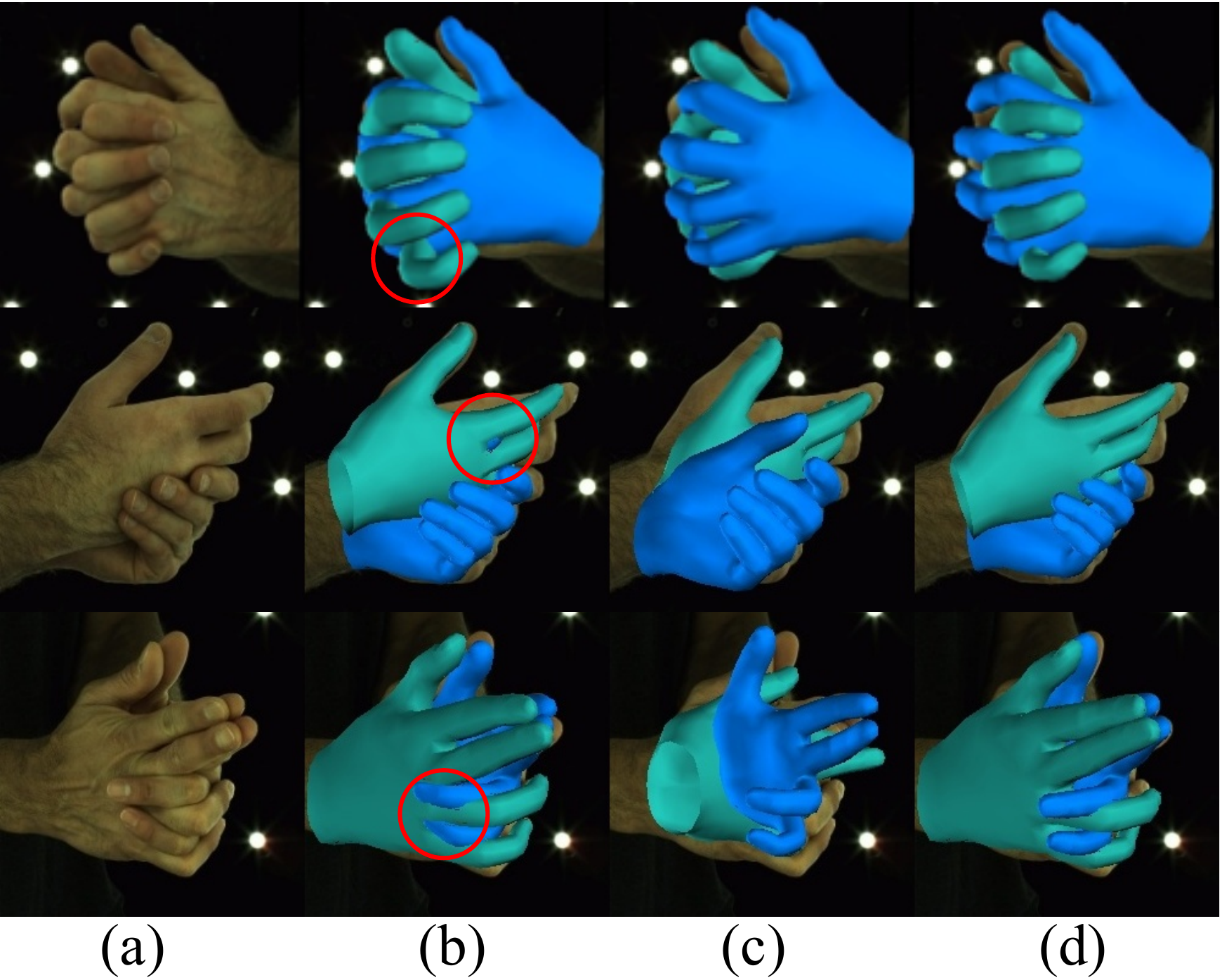}
	\end{center}
	\vskip -0.7cm
	\caption{\small \textbf{Interacting 3D Hands Reconstruction.} (a) Input RGB images with closely interacting hands. (b) A baseline model follows the single-hand based method~\cite{boukhayma20193d} to reconstruct two hands. The method tends to generate collided 3D hands. Collided regions are marked with red circles. (c) Results obtained by directly adding a collision-aware loss~\cite{jiang2020coherent} in training. (d) Results produced by the proposed method with mitigated collisions.}
	\label{fig:teaser}
	\vspace {-0.3cm}
\end{figure}

Capturing 3D interacting hand motion from monocular single RGB images can facilitate numerous downstream tasks including AR/VR~\cite{han2020megatrack,wang2021scene} and social signal understanding~\cite{joo2019towards,ng2020body2hands}. 
Previous works on motion capture of two interacting hands ~\cite{ballan2012motion,mueller_siggraph2019,oikonomidis2012tracking,taylor2017articulated,tzionas2016capturing,wang2020rgb2hands} mainly rely on depth images, multi-view images or image sequences as input. These methods cannot be easily applied to monocular RGB images.
Recently, there is a surge of interest to capture 3D single hand motion from monocular RGB images ~\cite{baek2019pushing,boukhayma20193d,ge20193d,kulon2020weakly,yang2020bihand,zhang2019end}.
For example, MANO~\cite{romero2017embodied}, a 3D hand mesh model parameterized by pose parameters and shape parameters, is proposed to represent 3D hands. Deep neural networks are leveraged to regress the model parameters or coordinates of the mesh vertices.
Although these methods achieve good performance in single-hand scenarios, it remains challenging to estimate interacting hands given just monocular RGB images. Typical results of a baseline model adapted from the single-hand methods~\cite{boukhayma20193d} are shown in Fig.~\ref{fig:teaser} (b). 
One major obstacle is that these methods do not explicitly model the relationship between two closely interacting hands, and therefore, inevitably generate unnatural results with hands colliding with each other.

An intuitive idea to diminish collisions is by applying collision-aware losses, inspired by previous works that model body-centric interactions~\cite{hassan2019resolving,jiang2020coherent,zhang2020perceiving}.
These losses are useful to increase the overall distances between interacting subjects, the main cause of collisions found across interacting bodies.
Collisions observed between interacting hands are attributed to multiple factors, such as imprecise hand shapes or finger poses.
Simply adding collision losses in model training inevitably makes a model converge to a trivial solution that merely pulls two hands away. 
While collisions are dismissed, the accuracy of 3D hand estimation is also compromised.. 
We illustrate several examples in Fig.~\ref{fig:teaser} (c).
A similar phenomenon has also been observed by previous body-motion studies, \eg, Jiang~\etal~\cite{jiang2020coherent}.

In this work, we introduce a two-stage framework that estimates 3D hand poses and shapes of two closely interacting hands with precise 3D poses and little collisions.
In the first stage, we adapt a convolutional neural network (CNN) with a similar architecture as single hand methods~\cite{boukhayma20193d,zhang2019end} to predict initial hand meshes of two hands.
Specifically, the CNN takes a single RGB image as input and regresses two-hand parameters as its output. 
In this stage, collision-aware losses are not adopted -- we deliberately allow the model to produce collided output but focus on generating acceptable and accurate pose results. 
These initial predictions provide good initialization for the second stage. Several examples are shown in Fig.~\ref{fig:teaser} (b).
From the coarse results obtained in the first stage, we observe that collisions of different samples are caused by distinct factors.
For example, the collisions shown in the first row of Fig.~\ref{fig:teaser} (b) are caused by inaccurate hand shape estimations while imprecise relative hand location causes the collisions in the second row. In the third row, collisions are mainly caused by the imperfect finger pose estimation.
An effective way to improve each collided initial prediction is to focus on its specific root cause of error instead of jointly optimizing for all components.

To this end, we present a novel factorized refinement strategy in the second stage to ameliorate the collision issue.
We decompose the causing factor of error and correct one factor at a time.
In this work, we consider different approaches to perform the factorized refinement, \ie, through the conventional gradient-descent based optimization and a more effective alternative that applies MLP as a proxy for optimization.
Using the proposed factorized refinement strategy, the collision issues can be substantially mitigated.
We show several recovered 3D hand meshes using the proposed two-stage framework in Fig.~\ref{fig:teaser} (d).
Compared with baseline results shown in Fig.~\ref{fig:teaser} (b) or results from a model trained with naively applying collision-aware losses, as shown in Fig.~\ref{fig:teaser} (c), our method can generate precise 3D hand meshes with modest collisions.

Our contributions can be summarized as follows: 
1) We introduce the first 3D interacting hand motion capture framework that only needs monocular single RGB images as inputs, without any other auxiliary sensory inputs.
2) We comprehensively investigate the collision issue that exists in interacting hands motion capture, and subsequently design a simple yet effective strategy to resolve the problem.
3) We present two alternatives for factorized refinement that carefully consider the trade-off in accuracy and speed. 
We perform extensive evaluations on the proposed method on large-scale interacting hand datasets, \eg, InterHand2.6M~\cite{moon2020interhand}.
In comparison to existing methods, the proposed approach achieves 71.4\% reduction in the generated collisions and improves pose estimation by 16.5\%.

%% file: sections/related_work.tex

\section{Related Work}

\newpara{3D Single Hand Pose Estimation.}
Early studies~\cite{oikonomidis2011efficient,sridhar2015fast,sridhar2013interactive,tkach2016sphere} take depth images as input and use optimization or discriminative methods to predict 3D joint locations. 
Subsequent works~\cite{ge2016robust,moon2018v2v,tompson2014real,wan2019self,ye2016spatial} replace the time-consuming optimization with deep neural networks.
Follow-up works~\cite{cai2018weakly,mueller2018ganerated,rong2021frankmocap,spurr2020weakly,yang2019disentangling,zimmermann2017learning} start to take RGB images as inputs and deploy CNN to estimate the 3D hand poses.
Inspired by recent progress in 3D body mesh recovery~\cite{kanazawa2018end,kolotouros2019spin,kolotouros2019cmr,rong2019delving,rong2020chasing,rong2021frankmocap,joo2021exemplar}, more recent approaches~\cite{baek2019pushing,boukhayma20193d,ge20193d,kulon2020weakly,yang2020bihand,zhang2019end} try to estimate 3D single-hand meshes from monocular RGB images.
Most works~\cite{baek2019pushing,boukhayma20193d,yang2020bihand,zhang2019end} use a CNN to predict the parameters of the MANO model~\cite{romero2017embodied}.
There are also other works directly regress the 3D vertices of the hand. Ge~\etal~\cite{ge20193d} and Pose2Mesh~\cite{choi2020pose2mesh} use graph neural networks. Kulon~\etal~\cite{kulon2020weakly} use mesh convolution. Lin~\etal~\cite{lin2021end} use Transformers~\cite{vaswani2017attention} as the backbone.
Although these methods perform well on single hands, it is non-trivial to adapt them to cope with interacting hands. 
Single-hand methods do not explicitly model the interrelationship between hands and thus will inevitably generate erroneous hand meshes, ~\eg hands colliding with each other.

\begin{figure*}[t]
	\begin{center}
		\includegraphics[width=0.95\linewidth]{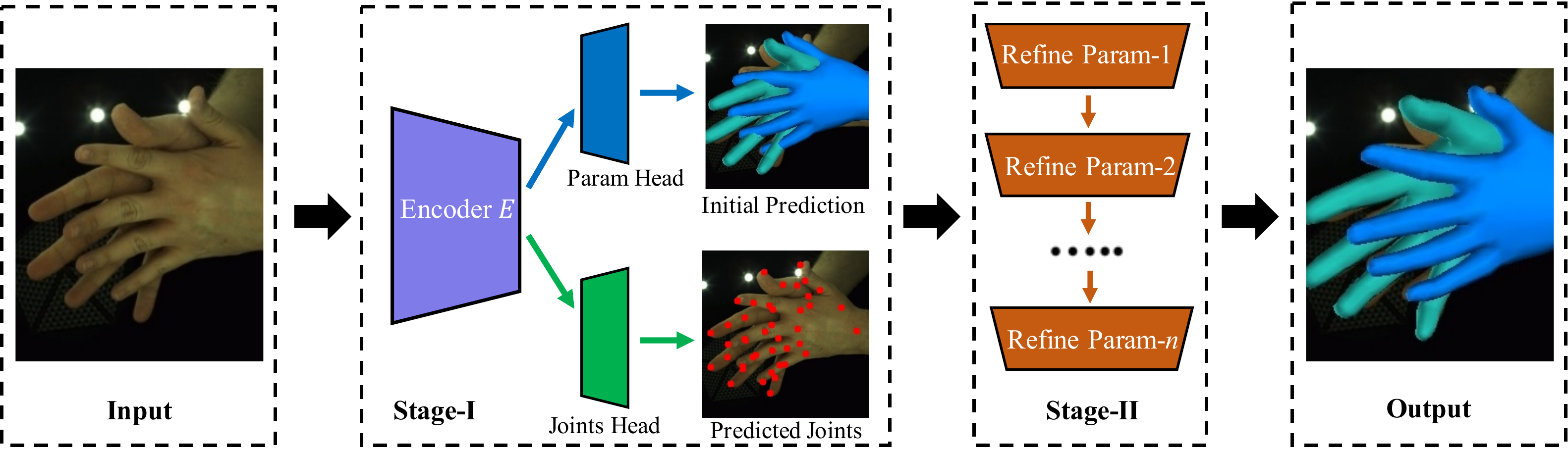}
	\end{center}
	\vskip -0.5cm
	\caption{\small \textbf{Framework overview.} The proposed \methodname~takes monocular RGB images as input and generates the 3D meshes represented in MANO~\cite{romero2017embodied}. The framework has two stages. The first stage adopts a convolutional neural network composed of an encoder, a MANO parameter prediction head, and a 2D/3D joint prediction head. It outputs the initial predicted parameters $\hbTheta$ and predicted 2D and 3D joints, namely $\dbj_{2D}$ and $\dbj_{3D}$. By taking initial predictions as input, the second stage performs factorized refinement to gradually remove the collisions from the first stage and produce refined predictions. The detailed architecture of each refinement module is depicted in Fig.~\ref{fig:refine}.}  
	\label{fig:framework}
	\vspace{-0.2cm}
\end{figure*}

\newpara{3D Interacting Hand Pose Estimation.}
Oikonomidis~\etal~\cite{oikonomidis2012tracking} firstly use Particle Swarm Optimization (PSO) to fit the 3D interacting hand motion that aligns best with the RGB-D inputs.
Ballan~\etal~\cite{ballan2012motion} solve this problem by estimating both finger-salient points and overall hand poses. They leverage the multi-view RGB sequences as the input.
Tzionas~\etal~\cite{tzionas2016capturing} extend the idea of discriminative salient points by introducing physics simulation.
Taylor~\etal~\cite{taylor2017articulated} introduce a new implicit hand model to facilitate simultaneous optimization over both hand poses and hand surfaces.
Smith~\etal~\cite{smith2020constraining} use an elastic volume deformation and a collision response model to recover dense hand surface from multi-view sequences.
%
%
More recent works of Muller~\etal~\cite{mueller_siggraph2019} and Wang~\etal~\cite{wang2020rgb2hands} share the similar pipeline. 
Specifically, deep neural networks are firstly used to predict the dense correspondence between pixels and MANO~\cite{romero2017embodied} surface. 
MANO parameters are then obtained by applying optimization to the obtained intermediate predictions.
Muller~\etal~\cite{mueller_siggraph2019} use single-view depth sequences while Wang~\etal~\cite{wang2020rgb2hands} use single-view RGB sequence as input.
Limited by the different input modality, these specialized methods cannot be directly applied to single RGB images.
InterNet~\cite{moon2020interhand}, although can cope with monocular RGB images, only produces 3D hand joint positions, which only have limited applications.
To fill the gap, we propose a new framework to simultaneously estimate both poses and shapes of interacting hands from single RGB images.

\newpara{Collision Detection across Interacting Hands.}
Oikonomidis~\etal~\cite{oikonomidis2012tracking} considers adjacent fingers interpenetrate by calculating abduction-adduction difference. 
Ballan~\etal~\cite{ballan2012motion} and Tzionas~\etal~\cite{tzionas2016capturing} use bounding volume hierarchies (BVH)~\cite{teschner2005collision} to detect each collided faces.
Muller~\etal~\cite{mueller_siggraph2019} and Wang~\etal~\cite{wang2020rgb2hands} use 3D Gaussians to model interpenetrate between fingers and palms.
Smith~\etal~\cite{smith2020constraining} does not only detect collided meshes but also adopt elasticity physical models to simulate the soft tissue of the hand skins.
In this paper, we use a modified Signed Distance Function (SDF)~\cite{hassan2019resolving,jiang2020coherent} which can model the penetration relations more precisely than finger-level~\cite{oikonomidis2012tracking}, BVH~\cite{ballan2012motion,tzionas2016capturing} or 3D Gaussians~\cite{mueller_siggraph2019,wang2020rgb2hands}. 
Besides, we propose to use the factorized refinement to effectively diminish the collision while retaining the accuracy of the estimated finger poses.

%% file: sections/method.tex

\section{Methodology}
\label{sec:method}

To reconstruct closely interacting 3D hands from monocular single RGB images, we propose a two-stage framework called Interacting Hand Mesh Recovery (\methodname).  
As depicted in Fig.~\ref{fig:framework}, in the first stage, a CNN similar to a previous single-hand model~\cite{boukhayma20193d} is adapted to generate initial predictions of two hands from input images. 
No collision-aware losses are applied in this stage so that the model can focus on generating accurate hand poses. 
A factorized refinement strategy is introduced in the second stage to gradually diminish the collisions observed in the first stage while retaining the accuracy of the estimated 3D hands.
We provide two different implementations for the proposed strategy, namely a optimization-based method with better accuracy and a neural-network based method with higher speed. 
Next, we first introduce the 3D hand model used in this paper. 
Then we discuss the design of the CNN baseline model and the factorized refinement strategy.
Lastly, we present the details of two different implementations of the latter.

\subsection {3D Hand Model}
\label{subsec:mano}

We use MANO~\cite{romero2017embodied}, a 3D hand mesh model parameterized by the shape and pose parameters.
The shape parameters $\bbetah \in \mathbb{R}^{10}$ control the overall shape of hands.
The pose parameters are composed of two parts. The first part $\bphih \in \mathbb{R}^{3}$ controls the global hand orientation.
The second part $\bthetah \in \mathbb{R}^{3 \times 15}$ represents the 3D rotations of each finger joint, relative to their parents on the predefined kinematics skeleton.
Hand orientation and finger poses are both represented in axis-angle format.
Given shape parameters $\bbetah$, hand orientation $\bphih$ and finger poses $\bthetah$, MANO model $W_h$ calculates the 3D vertices $\boldsymbol{V}_h \in \mathbb{R}^{3 \times 778}$ as:
\begin{equation}
	\label{eq:mano_origin}
	\boldsymbol{V}_h = W_h ( \bbetah, \bphih, \bthetah).\\
\end{equation}

The original MANO model only supports single hands. We extend it to support interacting hand scenarios by using two sets of parameters, namely $(\bbetal, \bphil, \bthetal)$ for left hands and $\left( \bbetar, \bphir, \bthetar \right)$ for right hands.
Another parameter $\btau \in \mathbb{R}^{3}$ is added to represent the relative translation from the right hand to the left one.
For simplicity of notation, we stack the same parameters of left and right hands together as $\bbeta = \left(\bbetal, \bbetar \right)$, $\bphi = \left(\bphil, \bphir \right)$, and $\btheta = \left(\bthetal, \bthetar \right)$.
The whole process of obtaining 3D interacting hand vertices from aforementioned parameters is formulated as:
\begin{equation}
	\begin{aligned}
	\label{eq:mano}
	\boldsymbol{V} &= W( \bbeta, \bphi, \btheta, \btau),\\
				   &= (W_{lh}(\bbetal, \bphil, \bthetal) + \btau, W_{rh}(\bbetar, \bphir, \bthetar)), \\
	\end{aligned}
\end{equation}
where $W_{lh}$ and $W_{rh}$ represent MANO models for left and right hands. 
$\bV = \left( \bV_l, \bV_r \right) $ is built from stacking left and right hand vertices.
3D joints $\boldsymbol{J}_{3D} \in \mathbb{R}^{3 \times 42}$ are obtained simultaneously in the process of linear blending skin.

\subsection{Basic 3D Hand Reconstruction Model}
\label{subsec:stage_one}

\newpara{Architecture.}
%
The baseline CNN model is composed of an encoder, a MANO parameter prediction head and a 2D/3D joint prediction head. 
Given single images as input, the encoder produces the encoded features $\bff$, which are then fed to the parameter prediction head. 
The parameter prediction head outputs the estimated parameters $\hbTheta = (\hbbeta, \hbphi, \hbtheta, \hbtau)$, as defined in Eq.~\eqref{eq:mano}.
Besides these parameters, the parameter head also predicts a set of weak-perspective camera parameters $\bpi \in \mathbb{R}^{3}$ to project the 3D joints into 2D space. 
The camera parameters $\bpi = (s, \boldsymbol{t})$ is composed of a scale factor $s$ and 2D translation $\boldsymbol{t} \in \mathbb{R}^{2}$.
Given camera parameters $\bpi$, the 2D joints $\bj_{2D}$ can be obtained as $\bj_{2D} = s\Pi(\bj_{3D}) + \boldsymbol{t}$, where $\Pi$ is the orthogonal projection.

%
%

\newpara{3D Losses.}
To train the baseline model, we adopt a set of 3D losses that include MANO parameter losses and 3D joint loss.
%
%
These losses are defined as follows:

\begin{equation}
	\label{eq:3d_loss}
	\begin{aligned}
		& L_{\Theta_h} = \lVert \abbeta - \hbbeta \rVert^{2}_{2} + \lVert R(\abtheta) - R(\hbtheta) \rVert^{2}_{2}, \\
		& L_{J3D} =  \lVert \abj_{3D} - \hbj_{3D} \rVert^{2}_{2}, \\  
		& L_{\tau} = \lVert \abtau - \hbtau \rVert^{2}_{2}, \\
		& L_{reg} = \lVert \hbbetal - \hbbetar \rVert^{2}_{2}, \\
	\end{aligned}
\end{equation}
where $\abbeta, \abtau, \abtheta, \abj_{3D}$ represent ground-truth MANO shape parameters, relative hand translation, finger poses and 3D joints. 
$\abtau$ is obtained from ground-truth 3D locations of left and right hand wrists.
$R$ is the Rodrigue's rotation formula, which converts the finger poses from the axis-angle format to the rotation-matrix format. We follow this practice of previous works~\cite{kanazawa2018end} for better numerical stability.
$L_{reg}$ is used to constrain the shape variation between two hands belonging to the same person.
We also tried surface-based 3D losses such as vertex coordinate loss, surface normal and edge losses used by Pose2Mesh~\cite{choi2020pose2mesh},
but empeircal results show that these losses bring no performance gain. 

\newpara{2D Losses.}
We also adopt a 2D loss defined as:
\begin{equation}
	\label{eq:2d_loss}
	\begin{aligned}
		& L_{J2D} = \lVert \mu \cdot (\abj_{2D} - \hbj_{2D}) \rVert_{1}, \\
	\end{aligned}
\end{equation}
where $\mu$ is the visibility indicator and $\abj_{2D}$ is the ground-truth 2D joints.

\newpara{Overall Loss.} The overall loss $L$ is defined as follows:
\begin{equation}
	\label{eq:overall_loss}
	\begin{aligned}
		L = & \lambda_{1}L_{\Theta_h} + \lambda_{2}L_{\tau} + \lambda_{3}L_{J3D} + \lambda_{4}L_{reg} + \lambda_{5}L_{J2D}
	\end{aligned}
\end{equation}
In our experiments, the values of these $\lambda$s are set to be $[10.0, 10.0, 10.0, 0.1, 10.0]$, respectively.

\subsection{Collision-Aware Factorized Refinements}
\label{subsec:factor_refine}
%

\subsubsection{Collision-aware Loss}
In this second stage, we adopt the penetration loss used by previous work~\cite{hassan2019resolving,jiang2020coherent} to reduce the collisions observed in the first stage.
Following their practice, we first use a modified Signed Distance Field (SDF) for each hand mesh, which is defined as follows:
\begin{equation}
	\label{eq:sdf}
	\begin{aligned}
		\psi_h(x,y,z) = -\min(\textrm{SDF}(x,y,z),0), \\
	\end{aligned}
\end{equation}
For points $(x,y,z)$ inside the hand meshes, the $\psi_h(x,y,z)$ takes positive values, which are in proportion to the points' distances to the hand surfaces. The values take zero for outside points.
The SDF is obtained via voxelizing the original 3D hand meshes to 3D grids with dimension $N_p \times N_p \times N_p$. 
We compute $\psi_l$ for left hands and sample the SDF value of each vertex of right hands, and vice versa. The collision loss is calculated as follows:

\begin{equation}
	\label{eq:collision_loss}
	\begin{aligned}
		L_{col} = \sum_{\boldsymbol{v} \in \boldsymbol{V}_l} \psi_r(\boldsymbol{v}) + 
		\sum_{\boldsymbol{v} \in \boldsymbol{V}_r} \psi_l(\boldsymbol{v}), \\
	\end{aligned}
\end{equation}
where $\psi_l$ and $\psi_r$ are SDF functions of left and right hand meshes, while $\boldsymbol{V}_l$ and $\boldsymbol{V}_r$ are left and right hands vertices.

Recall that we do not add the collision loss $L_{col}$ in training the baseline CNN model of the first stage. It is because adding collision-aware losses in model training will compromise the estimated 3D hands, as depicted in Fig.~\ref{fig:teaser} (c). 
Besides, even with a rigorous adjustment on the loss weight of $L_{col}$, we found that the model cannot converge to provide a balanced output, where pose estimation and collision status are both satisfactory. %
We verify this phenomenon in our experiments (Fig.~\ref{fig:collision_weight}), in which we list the baseline model's performances obtained by setting different loss weights of $L_{col}$. 
In view of this, we propose to have factorized refinement as a second stage to refine the outputs generated by the first-stage CNN model.

\begin{figure}[t]
	\begin{center}
		\includegraphics[width=1.0\linewidth]{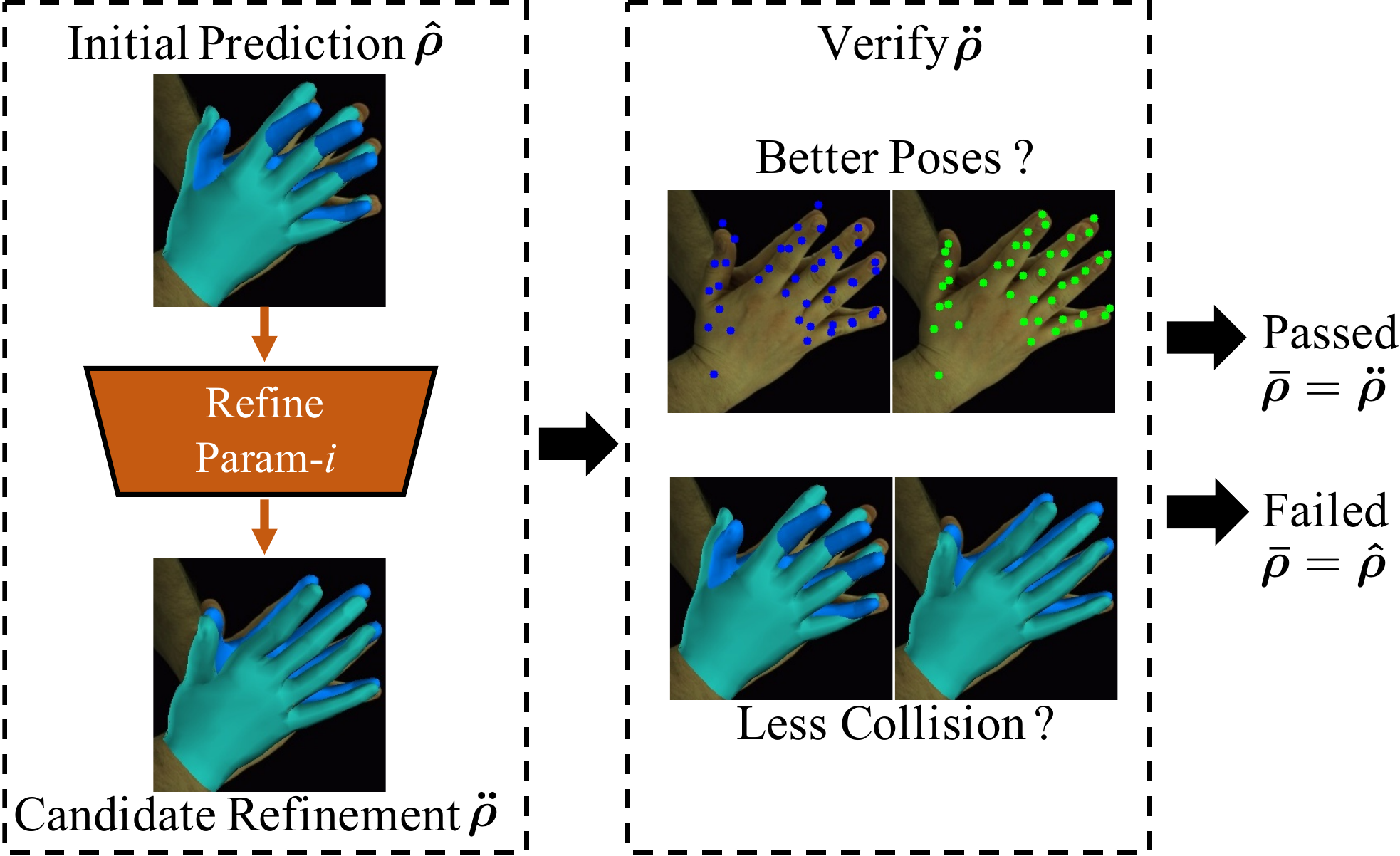}
	\end{center}
	\vskip -0.5cm
	\caption{\small \textbf{Refinement with verification.} Given original predictions generated by the baseline CNN model or previous refinement stages, the refinement module of current stage aims at refining the corresponding parameters and generating the candidate refinement. Some criteria are used to decide whether to keep the original predictions before refinement or adopt the refined parameters.}  
	\label{fig:refine}
	\vspace{-0.5cm}
\end{figure}

\subsubsection{Factorized Refinement}
We divide the cause factors of inter-hand collisions into four different parameter sets, namely imprecise right-to-left-hand translation $\hbtau$, imprecise hand orientations $\hbphi$, imprecise hand shapes $\hbbeta$ and imprecise hand finger poses $\hbtheta$. 
The goal of factorized refinement is to refine each cause factor given each collide sample.
However, due to the lack of corresponding annotations or explicit signals, it is challenging to use prior knowledge to precisely identify the major cause factors.
Therefore, we adopt stagewise refinement with verification, as shown in Stage-II of Fig.~\ref{fig:framework}, with each stage detailed in Fig.~\ref{fig:refine}. 
The whole process of factorized refinement is summarized in Algorithm~\ref{algo:refinement}.

Given each collided sample, we try to refine all four parameters in a stage-wise manner, following the order of $\hbrho \in \{\hbtau, \hbphi, \hbtheta, \hbbeta \}$.
The order is insignificant as we observed from our experiments.
We denote the initial predictions as $\hbrho$, the candidate obtained after the refinement module as $\tbrho$, and the refined parameters as $\bbrho$.
In each stage, we first refine one of the parameter sets, $\hbrho$, and obtain a candidate parameter set $\tbrho$. 
Then we check whether the candidate set could lead to better results, \eg, hand meshes with less collisions and lower 3D joint errors (detailed in next paragraph).
If the results are improved after verification, we use the candidate $\tbrho$ as the final refined parameters $\bbrho$. 
Otherwise, we still keep the original values $\hbrho$ as the output of this stage. 
Once we are done with the refinement of a parameter set we move to the next stage to refine another set of parameters.

For each refinement stage, we perform verification to decide if the candidate parameters could lead to a better performance. We adopt two criteria,~\ie, collision errors $\mE_{col}$ and 3D joint errors $\mE_{3D}$.
Collision errors can be calculated as Eq.~\eqref{eq:collision_loss} using the estimated vertices $\hbV$.
To calculate the 3D joint errors during inference time, we add another head onto the baseline CNN model to directly predict the 3D joints $\dbj_{3D}$, 
and use them as the pseudo-ground-truth 3D joints to calculate the 3D joint error as $\mE_{3D} = \lVert \dbj_{3D} - \hbj_{3D} \rVert^{2}_{2} $.
We follow the same architecture of the InterNet~\cite{sun2018integral} to use 3D heatmaps to build this additional joint prediction head. More details about the joint head are included in the supplemental material.
We use $\dbj_{3D}$ instead of regressed 3D joints from initial MANO predictions $\hbj_{3D}$ because it is revealed by the previous works that 3D joints predicted in the heatmap format have better accuracy than the joints regressed from mesh models~\cite{li2021hybrik}.
%

\input{sections/refine_algo.tex}


\newpara{Optimization-based Implementation.}
Our first implementation of stage-wise refinement uses gradient-descent based optimization, which updates the parameters by minimizing the following objective function:

\vskip -0.5cm
\begin{equation}
	\label{eq:opt_objective}
	\begin{aligned}
		\mF_{opt} = \mF_{col} + \mF_{2D} + 
			\mF_{3D} + \mF_{\tau} + \mF_{reg} + \mF_{f}, \\
	\end{aligned}
\end{equation}
where $\mF_{col}$ is computed using Eq.~\eqref{eq:collision_loss}.
The term $\mF_{2D}$ is computed in the same way as Eq.~\eqref{eq:2d_loss} by replacing the ground-truth 2D joints $\bj_{2D}$ with $\dbj_{2D}$, which is also estimated by the additional joint head used to predict $\dbj_{3D}$.
The term $\mF_{\tau}$ is computed in a similar way as $L_{\tau}$. The pseudo-ground-truth right-to-left-hand translation is obtained by subtracting the 3D left wrist location by the right one. The 3D joint locations are retrieved from $\dbj_{3D}$.
The term $\mF_{reg}$ is the same as $L_{reg}$ in Eq.~\eqref{eq:3d_loss}.

The last term $\mF_{f}$ is used to avoid generating twisted fingers during optimization. We follow previous study~\cite{zhang2019end} to formulate $\mF_{f}$. 
Let $\bp_a$, $\bp_b$, $\bp_c$, $\bp_d$ represent joints on the fingers from the tip to the palm. $\V_{ab} = \bp_b - \bp_a$ represents the finger vectors. There are two constraints listed as follows:
\begin{equation}
	\label{eq:finger_constrain}
	\begin{aligned}
		& \mathcal{C}_1 = (\V_{ab} \times \V_{bc}) \cdot \V_{cd} = 0 \\
		& \mathcal{C}_2 = (\V_{ab} \times \V_{bc}) \cdot (\V_{bc} \times \V_{cd}) \geqslant 0 \\
	\end{aligned}
\end{equation}
\noindent The finger error $\mF_f$ is defined as:
\begin{equation}
	\label{eq:finger_error}
	\begin{aligned}
		\mF_f = \lVert \mathcal{C}_1 \rVert + \min(\mathcal{C}_2, 0).
	\end{aligned}
\end{equation}
Detailed loss weights and learning rate used in each stage are listed in the supplemental.
At the end of each stage, verification criteria $\mE_{col}$ and $\mE_{3D}$ are used to select the valid updated parameters that can improve the performances.

\newpara{MLP-based Implementation.}
The optimization-based method offers a good performance but it runs at a slow speed ($0.9$ FPS). 
In view of this, we provide another implementation of factorized refinement using Multilayer Perceptrons (MLP), which can run in real time (29.6 FPS). 
Specifically, a separate MLP is deployed to refine each parameter set. The input of MLPs is the concatenation of encoded features $\bff$ generated by image encoder $E$ and the initial parameter predictions $(\hbbeta, \hbphi, \hbtheta, \hbtau)$. The output is the candidate parameters $\tbrho$, which is subjected to the verification step discussed earlier.
To train MLPs, we build a train subset consisting of only closely interacting samples. 
The training losses include 3D losses defined in Eq.~\eqref{eq:3d_loss}, 2D loss defined in Eq.~\eqref{eq:2d_loss} and collision loss defined in Eq.~\eqref{eq:collision_loss}.
%
Similar to the optimization-based implementation, criteria including $\mE_{col}$ and $\mE_{3D}$ are adopted during both train and test phases to filter out the candidate parameters that cannot lead to improved results.
Implementation details of the MLPs are included in the supplemental material.

%% file: sections/refine_algo.tex

\begin{algorithm}[t]
	\caption{Factorized Refinement Strategy.}
	\label{algo:refinement}
	
	\begin{algorithmic}[1] 
		\Require Initial parameters $\hbTheta = (\hbbeta, \hbphi, \hbtheta, \hbtau)$.
		\Require Pseudo-ground-truth 3D Joints $\dbj_{3D}$.
		\Require Initialize refined parameters $\bbTheta$ as $\hbTheta$.
		\Function {ObtainError}{$\bTheta, \dbj_{3D}$}
			\State $\bV = W(\bTheta) $ 
			\State $\bj_{3D} = ( J \cdot \bV^{T} )^T$
			\State $\mE_{col} = \Psi(\bV)$
			\State $\mE_{3D} =  \lVert \bj_{3D} - \dbj_{3D} \rVert^{2}_{2} $
			\State \Return $\mE_{col}, \, \mE_{3D}$
		\EndFunction
		
		\State $\bmE_{col}, \, \bmE_{3D} = \textrm{ObtainError}(\bbTheta, \dbj_{3D})$
		\For { $\bbrho$ in $\{ \bbbeta, \bbphi, \bbtheta, \bbtau \}$ }
			\State $\tbrho = \textrm{Refine}(\bbrho)$
			\State $\tbTheta = (\bbTheta \backslash \bbrho) \cup \{\tbrho\} $
			\State $\tmE_{col}, \tmE_{3D} = \textrm{ObtainError}(\tbTheta, \dbj_{3D})$
			\If{$\tmE_{col} < \bmE_{col} \,\,\textrm{and}\,\, \tmE_{3D} < \bmE_{3D} $}
				\State $\bbrho = \tbrho$
				\State $\bmE_{col} = \tmE_{col} $
				\State $\bmE_{3D} = \tmE_{3D} $
			\EndIf
		\EndFor
		
	\end{algorithmic}
\end{algorithm}
\setlength{\intextsep}{-10pt} 

%% file: sections/experiment.tex

\section{Experiments}


\subsection{Experimental Settings}
\label{subsec:exp_setting}


\newpara{Datasets.}
We majorly conduct experiments on InterHand2.6M~\cite{moon2020interhand}, which provides RGB images with annotated 2D and 3D joints.
The dataset also provides pseudo-ground-truth MANO~\cite{romero2017embodied} parameters obtained from NeuralAnnot~\cite{moon2020neuralannot} for part of samples.
Follow the practice of the original paper, we use the 5 FPS version of the released data.
Besides, we neglect the multi-view information and treat all data as single-view samples.
Our models are trained on the whole train set evaluated on the test set.
To focus our evaluation on the interacting hands, we select interacting samples from the whole test set.
Following the same definition as the original paper, we attribute samples with more than 30 valid ground-truth 3D joints as ``interacting''.
Besides the interacting subset, another subset composed of only \textbf{closely} interacting samples is further filtered out.
Given each interacting sample, we first calculate the per-joint ``inter-distance'', which is defined as the smallest Euclidean distance from this joint to the joints belonging to the other hand.
Closely interacting samples are defined as the sample whose average inter-distance of all valid joints is smaller than 40 mm.
We simplify the names of the original test set, the interacting subset, and the closely interacting subset as ``IH26M-All'', ``IH26M-Inter'', and ``IH26M-Inter-Close''.
%
%
We also train and evaluate our models on RGB2Hand dataset~\cite{wang2020rgb2hands} and Tzionas dataset~\cite{tzionas2016capturing}.
We follow the same practice of InterNet~\cite{moon2020interhand} to randomly split the origin Tzionas dataset~\cite{tzionas2016capturing} into train and test set at a ratio of 9:1 and additionally train and evaluate our models on the split sets.

\newpara{Evaluation Metrics.}
We first adopt a single-hand style metric, 3D Mean Per Joint Position Error (MPJPE).
In calculating MPJPE, the left and right hand joints are separately aligned to the ground-truth left and right wrists.
We adopt another interacting-hand style metric I-MPJPE that synchronically aligns both the left and right hands joints to the ground-truth joints using Procrustes Analysis.
Following the practice of RGB2Hands~\cite{wang2020rgb2hands}, we only normalize the scales and translations while leaving the estimated rotations unchanged.
Furthermore, we adopt two more metrics to reveal the collision status of the reconstructed interacting hands. 
They are Average Penetration Depth (AVE-P) of each hand vertices penetrating into the other hand, and Maximum Penetration Depth (MAX-P) of all vertices belonging to a pair of hand. 
In the last, we evaluate samples with pseudo-ground-truth MANO annotations using Mean Per Vertex Position Error (MPVPE) and corresponding I-MPVPE.
These results are listed in the supplemental.
Units of the aforementioned metrics are all millimeters (mm).

\subsection{Comparison}

\input{tables/compare_sota.tex}

\begin{figure*}[t]
	\begin{center}
		\includegraphics[width=0.9\linewidth]{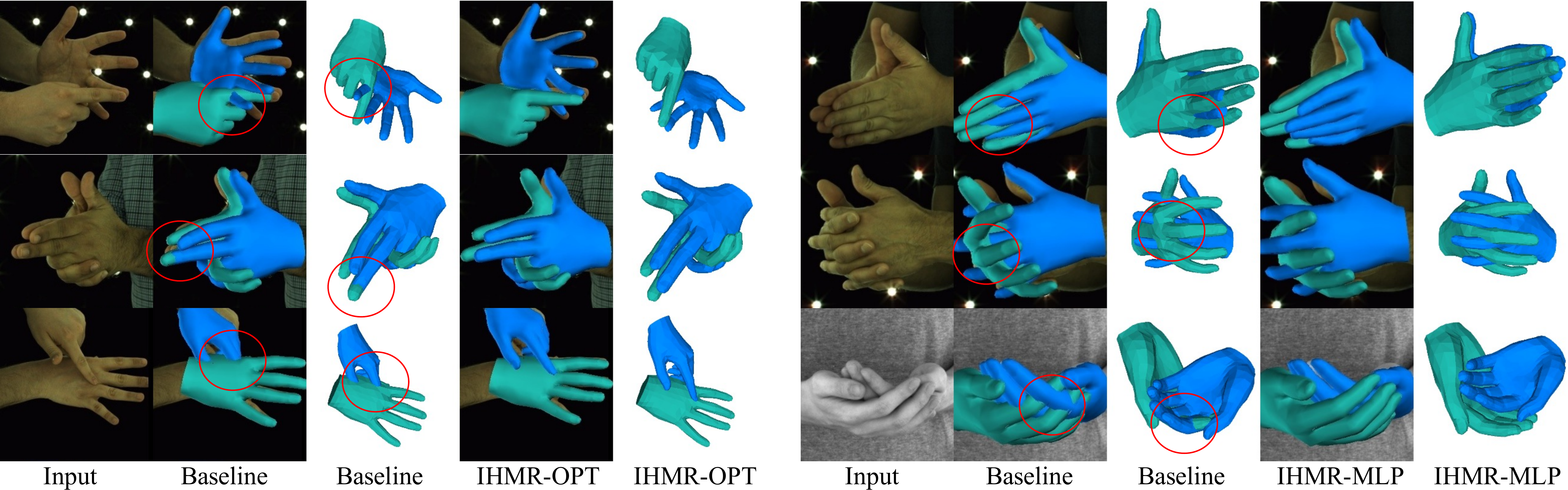}
	\end{center}
	\vskip -0.6cm
	\caption{\small \textbf{Qualitative results on InterHand2.6M}~\cite{moon2020interhand}. Quanlitative comparison between baseline model adopted from Bouk~\etal~\cite{boukhayma20193d} and \methodname. Collided regions are marked with red circles.}  
	\label{fig:qualitative}
	\vspace{-0.3cm}
\end{figure*}

\newpara{Comparison on InterHand2.6M.}
The quantitative results on InterHand2.6M~\cite{moon2020interhand} are listed in Tab.~\ref{tab:compare_sota}.
We first compare performances of three single-hand methods (Bouk~\etal~\cite{boukhayma20193d}, Pose2Mesh~\cite{choi2020pose2mesh} and BiHand\cite{yang2020bihand}) on test set of InterHand2.6M using single-hand metric MPJPE.
All these three methods are finetuned on the train set of InterHand2.6M.
As the models' performances are close, we choose Bouk~\etal~\cite{boukhayma20193d} as the baseline model to build up~\methodname~for its simplicity.

Due to the lack of similar works, we majorly compare the variants of~\methodname.
They are the baseline CNN model (\methodname-Baseline), \methodname~with factorized refinement implemented in MLP (\methodname-MLP), and~\methodname~implemented in optimization (\methodname-OPT).
We also test the models' inference speed on a single RTX 2080Ti GPU.
The results show that the proposed factorized refinement strategy can effectively diminish the collisions between the interacting hands while producing more precise 3D poses.
On the most challenging IH26M-Inter-Close,~\methodname-OPT reduces the AVE-P by 71.4\% while improving the interacting pose estimation accuracy by (in terms of I-MPJPE) by 16.5\%.
On the other hand,~\methodname-MLP can reduce the collision and 3D finger poses errors by 26\% and 10.4\%.
We believe the performance gap between~\methodname-MLP and~\methodname-OPT is caused by their inherent differences. 
Optimization is better at modeling high-frequency information such as per-vertex collisions.
On the other hand,~\methodname-MLP achieves real-time performance, which is more suitable for applications such as live streaming that demand real-time procesing.

Furthermore, we adopt HybrIK~\cite{li2021hybrik} as an additional strong baseline to fill the gap between InterNet~\cite{moon2020interhand}.
Implementation details are included in the supplemental.
Based on this strong baseline, we further adopt factorized refinement. 
The results are listed in the last $4$th to $2$nd rows (marked with $\boldsymbol{^{*}}$) of Tab.~\ref{tab:compare_sota}.
By following HybrIK, the I-MPJPE gap between InterNet and~\methodname-Baseline$\boldsymbol{^{*}}$ is reduced from $8.3$ mm to $3.5$ mm.
Furthermore, the proposed factorized refinement can still effectively increase the accuracy of 3D joints estimation and reduce collisions despite the stronger baseline.
Note that our method is advantageous than InterNet in that our reconstructed 3D meshes allow for more elaborative interactions such as "touching" and "grabbing" compared to 3D joints produced by InterNet.

Qualitative results are shown in Fig.~\ref{fig:qualitative}. 
It reveals that \methodname, no matter implemented using the MLP or the optimization, can effectively diminish the collisions caused by different factors,~\eg relative hand positions (the first row on the left half) or finger poses (the second row on the right half).
See the supplemental for more qualitative results, including the comparison with single-hand methods and the comparison between \methodname-MLP and \methodname-OPT.

\begin{figure}[t]
	\begin{center}
		\includegraphics[width=0.8\linewidth]{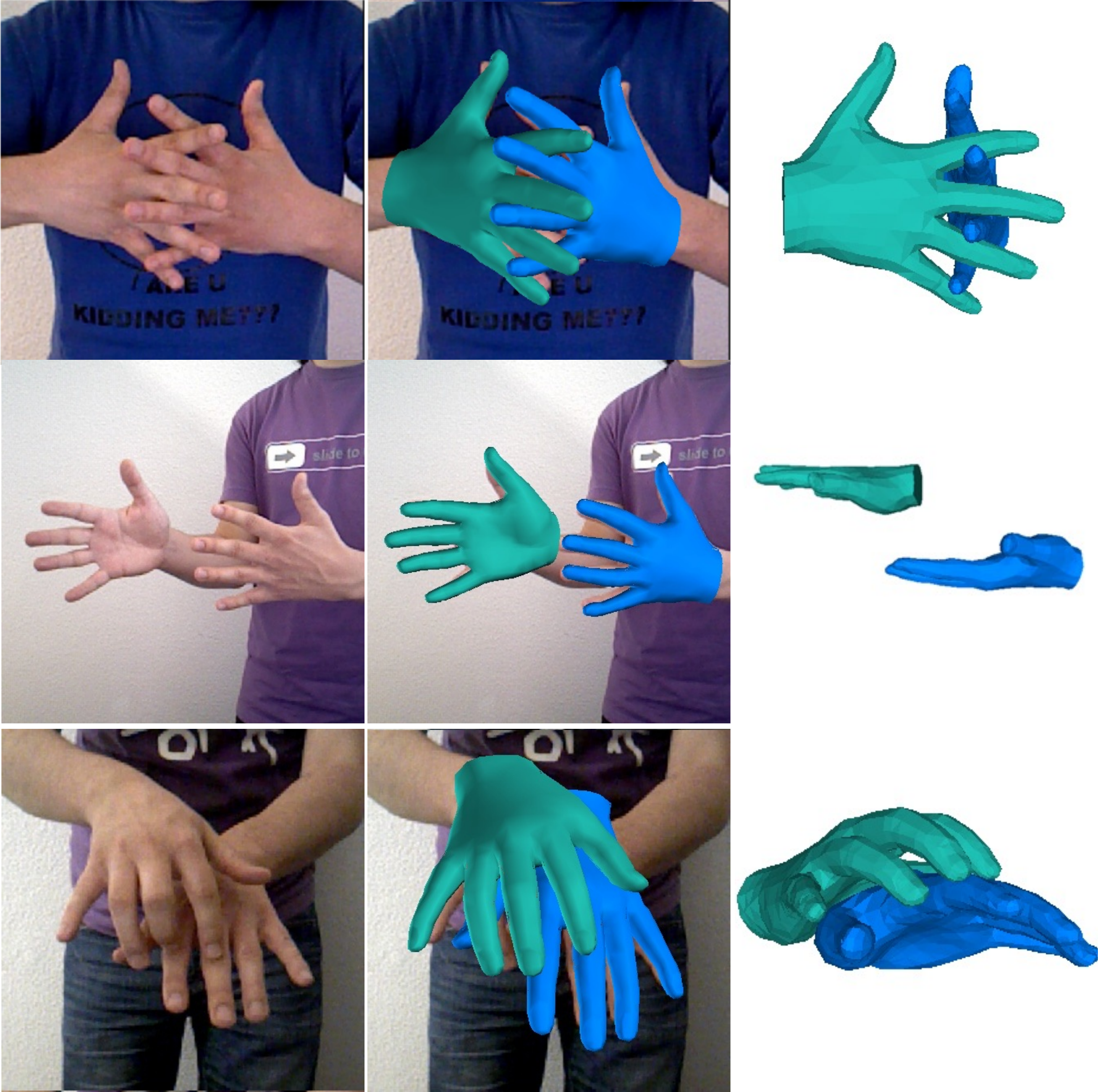}
	\end{center}
	\vskip -0.6cm
	\caption{\small \textbf{Qualitative results on Tzionas dataset}~\cite{tzionas2016capturing}.}
	\label{fig:tziona_qualitative}
	\vspace{-0.7cm}
\end{figure}

\input{tables/compare_rgb2hand.tex}

\newpara{Comparisons on Other Datasets.}
We quantitatively compare our methods with RGB2Hands~\cite{wang2020rgb2hands} on the RGB2Hands evaluation set. 
The results are listed in Tab~\ref{tab:compare_rgb2hand}. RGB2Hands perform slightly better than our methods.
We believe it is due to that RGB2Hands leverage temporal information while we only use single RGB images as inputs.
For Tzionas dataset~\cite{tzionas2016capturing}, we only show qualitative results in Fig.~\ref{fig:tziona_qualitative} due to the lack of 3D annotations.
We follow the practice of InterNet~\cite{moon2020interhand} to train the model with the mixture of InterHand2.6M data and Tzionas data.
These results demonstrate the \methodname's generalization ability.

\subsection{Ablation Study}
Models in this subsection are based on original~\methodname~described in Sec.~\ref{sec:method} and evaluated on IH26M-Close-Inter, using I-MPJPE and AVE-P as the metrics.
More ablation studies are included in the supplemental.

\begin{figure}[t]
	\begin{center}
		\includegraphics[width=1.0\linewidth]{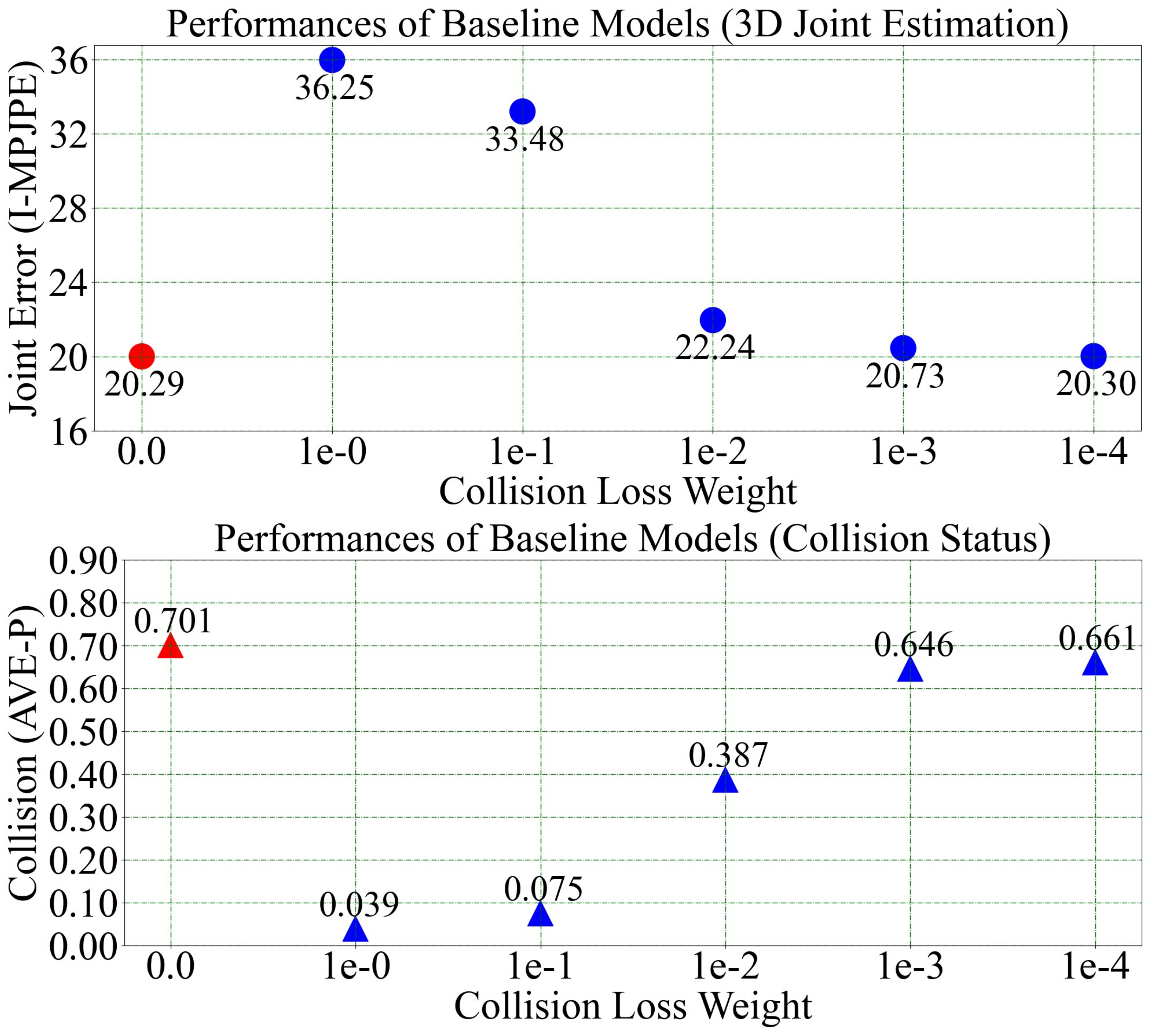}
	\end{center}
	\vskip -0.65cm
	\caption{\small \textbf{Collision loss in training baseline CNN model.} We apply collision loss $L_{col}$ in training the baseline CNN model. Red markers represent the original model trained without collision loss.}  
	\label{fig:collision_weight}
	\vspace{-0.35cm}
\end{figure}

\newpara{Adopting Collision Loss in Stage-I.}
We conduct experiments in which the collision loss $L_{col}$ in Eq.~\eqref{eq:collision_loss} is directly applied in training the baseline CNN model. 
We adjust the loss weights of $L_{col}$ to range from $0.0$ to $1\mathrm{e}{-4}$, where $0.0$ corresponds to the original baseline model.
All models in this subsection are finetuned from a model trained on InterHand2.6M without collision losses adopted.
The results are depicted in Fig.~\ref{fig:collision_weight}. It is observed that although using collision loss in training baseline model can remove most of the collisions, it corrupts the 3D pose estimation, with the joint estimation errors increased by 79\%.
Furthermore, simply adjusting the loss weight of $L_{col}$ cannot lead to better results with 3D hand pose estimation and collision status both improved.
While decreasing loss weights of $L_{col}$ can restrain the performance drop in the joint estimation, the collision issue becomes severe again.
The results justify our design of not using collision-aware loss in the first stage but introducing it in the second stage using factorized refinement.

\input{tables/ablation_strategy.tex}
\newpara{Design of Factorized Refinement.}
We evaluate three variants of ~\methodname. 
The first one is the framework without factorization, namely all parameters are refined together. 
The second one is performing factorized optimization without verification. 
The third one optimizes from mean hand poses, following the classical optimization methods such as SMPLify-X~\cite{pavlakos2019expressive}.
The results are listed in Tab.~\ref{tab:ablation_strategy}. 
Without verification, both methods fail - the 3D joint errors of~\methodname-MLP increase by 47\% and the average penetration of~\methodname-OPT increases by 25\% percent.
Without factorization, the average penetration of~\methodname-MLP increases by 19\%.
The gain is less apparent on~\methodname-OPT due to its unique per-sample optimization capability, which allows~\methodname-OPT to achieve optimal results more easily. 
Lastly, optimization starting from mean poses produces results with 3D joint errors 41\% larger than~\methodname-OPT does, which justifies the design of our two-stage framework.

\input{tables/ablation_param.tex}
\newpara{Influence of Refining Different Parameters.}
Studies on the influence of refining different parameter combinations are listed in Tab.~\ref{tab:ablation_param}.
Solely optimizing shape parameters provides the least performance gain. Updating finger poses can reduce the most 3D pose estimation errors, while refining orientation can diminish the most collisions.
Besides, the performances can be further improved by refining more than one parameter in the second stage.
Starting from the shape parameters, progressively incorporating orientation, finger poses, and translation in the whole process can consistently improve the overall performances.

%% file: tables/compare_sota.tex

\begin{table*}[t]
	\begin{center}
		\caption{\small \textbf{Comparison with SOTA methods on InterHand2.6M}~\cite{moon2020interhand}. We apply MPJPE and I-MPJPE to evaluate 3D joint estimation accuracy in the single-hand style and interacting-hand style. AVE-P and MAX-P are adopted to estimate the collision status of the generated interacting hands. We also test the speed of the variants of~\methodname, the unit is frame-per-second (FPS).}
		\vskip -0.3cm
		\resizebox{2.0\columnwidth}{!}{
			\begin{tabular}{l|c|ccc|ccc|c}
				\hline
				Dataset \& Metrics $\rightarrow$   & IH26M  & \multicolumn{3}{c|}{IH26M-Inter} & \multicolumn{3}{c|}{IH26M-Inter-Close}  & Speed  \\
				\hline
				Methods $\downarrow$ 			   & MPJPE $\downarrow$ & MPJPE $\downarrow$ & I-MPJPE $\downarrow$ & AVE-P / MAX-P $\downarrow$  & MPJPE $\downarrow$ & I-MPJPE $\downarrow$  & AVE-P / MAX-P $\downarrow$ & Speed (FPS) $\uparrow$ \\
                \hline
                Bouk~\etal~\cite{boukhayma20193d}  & 27.14   & 31.46   & -  &  -  & 26.64 & -  & - & - \\
			    Pose2Mesh~\cite{choi2020pose2mesh} & 27.10   & 32.11   & -  &  -  & 25.97 & -  & - & - \\
                BiHand~\cite{yang2020bihand}       & 25.10   & 28.23   & -  &  -  & 24.16 & -  & - & - \\
				\hline
				\methodname-Baseline 	& 25.45    & 29.70   & 39.38  & 0.47 / 10.65  & 26.30 & 20.29 & 0.70 / 14.10  & \tb{76.8} \\
				\methodname-MLP      	& 24.51    & 28.72 	 & 32.75  & 0.32 / 8.34   & 24.74 & 18.18 & 0.52 / 12.00  & 29.6 \\
				\methodname-OPT 		& 22.88    & 29.57   & 23.82  & \tb{0.09} / \tb{3.15}   & 24.61 & 16.94 & \tb{0.20} / \tb{6.23} & 0.89 \\
				\hline
				\methodname-Baseline$^{*}$ & 18.52 & 21.78   & 20.03  & 0.27 / 5.53   & 19.63 & 15.35 & 0.64 / 10.35 & 75.2 \\
				\methodname-MLP$^{*}$      & 17.61 & 21.23   & 19.45  & 0.25 / 5.34   & 18.60 & 14.42 & 0.44 / 9.55 & 29.6 \\
				\methodname-OPT$^{*}$      & \tb{17.12} & \tb{20.66}   & \tb{19.05}  & 0.21 / 4.48   & \tb{17.66} & \tb{13.56} & 0.27 / 6.27 & 0.89 \\   
				\hline	 
				InterNet~\cite{moon2020interhand}        & 14.21  & 18.04  & 17.36  &  -  & 16.19    & 11.91   & - & - \\
				\hline
			\end{tabular}
		}
		\vspace {-0.4cm}
		\label{tab:compare_sota}
	\end{center}
\end{table*}

%% file: tables/compare_rgb2hand.tex

\begin{table}[t] \centering 
	\caption{\small Comparison with RGB2Hands~\cite{wang2020rgb2hands} on their test set.}
	\vskip -0.3 cm
	\resizebox{1.0\columnwidth}{!}{
		\begin{tabular}{c|ccc}
			\hline
			Method & RGB2Hands~\cite{wang2020rgb2hands} & IHMR-MLP & IHMR-OPT \\
			\hline
			I-MPJPE $\downarrow$ & \textbf{20.02} & 23.41 & 21.30 \\
			\hline
		\end{tabular}	
	}
	\vspace{-0.5 cm}
	\label{tab:compare_rgb2hand}
\end{table}

%% file: tables/ablation_strategy.tex

%
\begin{table}[t] \centering 
	\caption{\small \textbf{Design of factorized refinement.} We examine three variants of the ~\methodname: without factorization and without verification, and optimization from mean hand poses}
	\vskip -0.3cm
	\setlength\tabcolsep{4 pt}
	\resizebox{0.8\columnwidth}{!}{
		\begin{tabular}{c|cccc}
			\hline 
			\mrow{Models $\downarrow$} & \multicolumn{2}{c}{I-MPJPE $\downarrow$} & \multicolumn{2}{c}{AVE-P $\downarrow$} \\
			& OPT & MLP & OPT & MLP \\
			\hline
			Full Model		 & 16.94 & 18.18 & 0.20 & 0.52 \\
			No Factorization & 17.20 & 18.98 & 0.22 & 0.62 \\
			No Verification  & 16.19 & 26.64 & 0.25 & 0.22 \\
			OPT Mean Pose	 & 34.74 & 26.22 & 0.03	& 3.67 \\
			\hline 
		\end{tabular}	
	}
	\vspace{-0.3cm}
	\label{tab:ablation_strategy}
\end{table}

%% file: tables/ablation_param.tex

%
\begin{table}[t] \centering 
	\caption{\small \textbf{Influence of refining different parameters.} We evaluate shape parameters $\bbeta$, global hand orientation $\bphi$, finger poses $\btheta$ and right-to-left-hand relative translation $\btau$. Full factorized refinement with verification is applied in these experiments.}
	\vskip -0.3cm
	\setlength\tabcolsep{3 pt}
	\resizebox{\columnwidth}{!}{
		\begin{tabular}{cccc|cccc}
			\hline 
			\mrow{Shape} & \mrow{Orientation} & \mrow{Finger} & \mrow{Translation} & \multicolumn{2}{c}{I-MPJPE $\downarrow$} & \multicolumn{2}{c}{AVE-P $\downarrow$} \\
			& & & & OPT & MLP & OPT & MLP \\
			\hline
			\xmark & \xmark & \xmark & \xmark	    & 20.29 & 20.29 & 0.70 & 0.70 \\ 
			\hline
			\cmark   & & & 	             			& 20.14 & 19.43 & 0.60 & 0.72 \\
			& \cmark & &                			& 19.18 & 19.29 & 0.41 & 0.64 \\
			& & \cmark &                 			& 19.28 & 19.42 & 0.40 & 0.71 \\
			& & & \cmark                 			& 18.67 & 19.32 & 0.51 & 0.67 \\
			\hline
			\cmark & \cmark & &          			& 19.16 & 19.43 & 0.51 & 0.65 \\
			\cmark & \cmark & \cmark &   			& 18.38 & 19.42 & 0.38 & 0.63 \\
			\cmark & \cmark & \cmark & \cmark 		& \textbf{16.94} & \textbf{18.18} & \textbf{0.20} & \textbf{0.52} \\
			\hline 
		\end{tabular}	
	}
	\vspace{-0.3cm}
	\label{tab:ablation_param}
\end{table}

%% file: sections/conclusion.tex

\section{Conclusion}

We have presented the first monocular single-image based method for reconstructing 3D interacting hands with both accurate finger poses and moderate inter-collisions.
To achieve this, we have proposed a novel factorized refinement that can be implemented in either traditional optimization or a feed-forward MLP. 
Extensive experiments on large-scale datasets demonstrate the effectiveness of our approach, which can consistently generate better 3D hand predictions despite the strong baseline such as HybrIK.

\noindent \textbf{Acknowledgement.} This study is supported under the RIE2020 Industry Alignment Fund Industry Collaboration Projects (IAF-ICP) Funding Initiative, as well as cash and in-kind contribution from the industry partner(s). It is also partially supported by the NTU NAP grant.
%
%
%

%
%

%

%% file: sections/supp.tex


\clearpage
\appendix

{\noindent\Large\textbf{Appendix}}
\counterwithin{figure}{section}
\counterwithin{table}{section}

\section{Implementation Details}

In this section, we introduce the implementation details of ~\methodname. 
We first introduce data preprocessing.
Then we introduce the architecture and training details of the baseline CNN model used in stage-I. 
After that, we introduce the design details of the optimization. 
In the end, we introduce the architecture and training details of the MLP-based implementation.

\subsection{Data Preprocessing}
We tightly crop the input images surrounding the bounding boxes of the interacting hands.
The hand bounding boxes are obtained from the 2D keypoint annotations.
For InterHand2.6M dataset~\cite{moon2020interhand}, we assume the known of hand types,~\ie single hand (left or right) or interacting hands.
To increase the model's generalization ability to in-the-wild scenarios, where no such labels are provided, we follow the practice of InterNet~\cite{moon2020interhand} to additionally predict the hand type (left, right, or interacting) from input images.
The trained model achieved 96\% prediction accuracy on the whole test set of InterHand2.6M.

\subsection{Baseline CNN Model}
The input images are cropped, padded, and resized to $224 \times 224$. 
The encoder is a ResNet-50~\cite{he2016deep}. 
The parameter prediction head is composed of three fully connected layers whose output dimensions are $1024$, $1024$, and $122$.
The $122$ dimensions are composed of camera parameters ($\bpi \in \mathbb{R}^{3}$), shape parameters ($\bbeta = (\bbeta_l,\bbeta_r) \in \mathbb{R}^{20}$), hand orientation ($\bphi = (\bphi_l, \bphi_r) \in \mathbb{R}^{6}$) finger pose parameters ($\btheta = (\btheta_l,\btheta_r) \in \mathbb{R}^{90}$), and right-to-left-hand relative translation ($\btau \in \mathbb{R}^{3}$).
The joint head follows the design as InterNet~\cite{moon2020interhand}.
The whole framework, including the encoder, the parameter prediction head, and the joint head is trained end-to-end using Adam optimizer~\cite{kingma2015adam} with a learning rate $1\mathrm{e}{-3}$. The whole model converges after $20$ epochs.

\subsection{Optimization-based Implementation}
The optimization-based method uses Adam optimizer~\cite{kingma2015adam} to directly update the estimated parameters $\hbrho \in \{\hbtau, \hbphi, \hbtheta, \hbbeta \}$. When refining each parameter set, we use different step size $\gamma$ and different weights for each objective listed in Equ.(8) and of the main paper.
The values of step size and objective weights are listed in Tab.~\ref{tab:opt_implement}.
We set smaller objective weights and step sizes in optimizing right-to-left-hand relative translation $~\hbtau$. Otherwise, the optimization tends to unreasonably increase the overall distances between collided hands and thus generate worse results.

\subsection{MLP-based Implementation}
In the MLP-based implementation, each refinement stage is composed of a Multilayer Perceptron with four fully connected layers, whose output dimensions are $512$, $256$, $128$, and $K$, where $K$ is the dimensions of the corresponding parameters.
The dimensions for each parameter are $3$ for hand translation $\hbtau$, $6$ for hand orientation $\hbphi$, $20$ for shape parameters $\hbbeta$ and $90$ for finger poses $\bbtheta$.
The training set of these MLPs are composed of samples with closely interacting hands, following the same selecting criteria of ``IH26M-Inter-Close`` defined in Sec.4.1 of the main paper.
We use the \methodname-OPT to obtain pseudo-ground-truth MANO parameters for train samples without MANO annotations.
The loss weights for training the MLPs are the same as Equ.(5) of the main paper.
Each MLP is trained for $2$ epochs with the learning rate set as $1\mathrm{e}{-4}$. The optimizer is Adam~\cite{kingma2015adam}.

\subsection{HybrIK Implementation}
To fill the performance gap between InterNet~\cite{moon2020interhand}, we adopt a strong baseline following HybrIK~\cite{li2021hybrik}.
We use the estimated 3D joints $\dbj_{3D}$ from the joint head of Stage-\Romannum{1}.
To be specific, the relative right-to-left-hand translation $\hbtau$ is directly set as the subtraction of predicted left wrist joints by the right one.
The global hand orientation $\hbphi$ is calculated using the locations of wrists and the five palm joints through Singular Value Decomposition (SVD).
The finger poses $\hbtheta$ are calculated in the standard way of HybrIK.
Please be noted that finger rotations are only composed of swing rotations.
Therefore, the finger rotations can be directly solved from finger joint locations.
In the last, we use the original shape parameters $\hbbeta$ predicted from the baseline CNN model of Stage-\Romannum{1}.
We suggest to read the original paper of HybrIK~\cite{li2021hybrik} to have a better understanding of how HybrIK is adopted in our scenario.

\input{tables/opt_implement.tex}
%
\input{tables/compare_sota_mpvpe.tex}
%
\input{tables/ablation_objective.tex}
%
\input{tables/ablation_joint.tex}

\section{More Quantitative Results.}
Part of the training and testing data have pseudo-ground-truth MANO~\cite{romero2017embodied} parameters obtained from NeuralAnnot~\cite{moon2020neuralannot}. 
The number of samples with pseudo GT MANO annotations are $240$K, $65$K, $12$K, and $4.5$K for all four subsets.
For samples with pseduo GT MANO annotations, We further calculate Mean Per Vertex Position Error (MPVPE) and I-MPVPE to reveal the quality of estimated joint rotations and shapes.  
They follow a similar definition as MPJPE and I-MPJPE.
The results are included in Tab.~\ref{tab:compare_sota_mpvpe}.
The conclusion we draw from Tab.~\ref{tab:compare_sota_mpvpe} is similar to the conclusion we draw from Tab.1 of the main paper.
The baseline models have similar performances.
On the most challenging IH26M-Inter-Close test set,~\methodname-OPT reduces the AVE-P by 60.7\% while improving the accuracy of interacting pose estimation and 3D hand reconstruction by 14.3\% and 16.0\%.
On the other hand,~\methodname-MLP can reduce the collision, 3D finger poses error and 3D hand reconstruction error by 23.5\%, 9.0\%, and 9.7\%.

\section{More Ablation Studies}
Following the same practice of the main paper, 
models in this subsection are evaluated on IH26M-Close-Inter, using I-MPJPE and AVE-P as the metrics.
More ablation studies are included in the supplemental.
\subsection{Influence of Different Optimization Objectives.}
In this subsection, we study the role of several objectives used in the optimization implementation of the factorized refinement.
The studied objectives include the 2D objective $\mF_{2D}$, the 3D objective $\mF_{3D}$ and the translation objective $\mF_{\tau}$. The results are listed in Tab.~\ref{tab:ablation_objective}.
When 2D or 3D joint objectives are adopted, both the 3D joint estimation error and collision status can be reduced.
When there is only the translation objective been adopted, although the collisions are almost totally removed, the joint estimation is ruined.
In general, adopting all three objectives can lead to be best result with both good joint pose estimation and less collision status.

\subsection{Influence of Joint Quality.}
To evaluate the convergence of the proposed factorized refinement, we conduct experiments in which pseudo-ground-truth 3D joints $\dbj_{3D}$ are replaced with ground-truth 3D joints with noise.
The noise is sampled from Gaussian distribution with $0$ mm mean and standard deviation ranging from $0$ mm to $40$ mm. 
For 2D joints, we still use the original pseudo-ground-truth 2D joints $\dbj_{2D}$ obtained from the joint head.
The results are listed in Tab.
It is revealed that: (1) The proposed factorized refinement is robust to the noise. 
When the standard deviation of the added noise is increased to $40$ mm,~\methodname-OPT and ~\methodname-IHMR can still decrease the 3D joint estimation error by 17.6\% and 5.7\%.
(2) The accuracy of 3D joint estimation is in proportion to the preciseness of the pseudo ground-truth 3D joints $\dbj_{3D}$.
(3) The effectiveness of collision removal has less correlation with the quality of $\dbj_{3D}$.

\begin{figure*}[t]
	\begin{center}
		\includegraphics[width=0.8\linewidth]{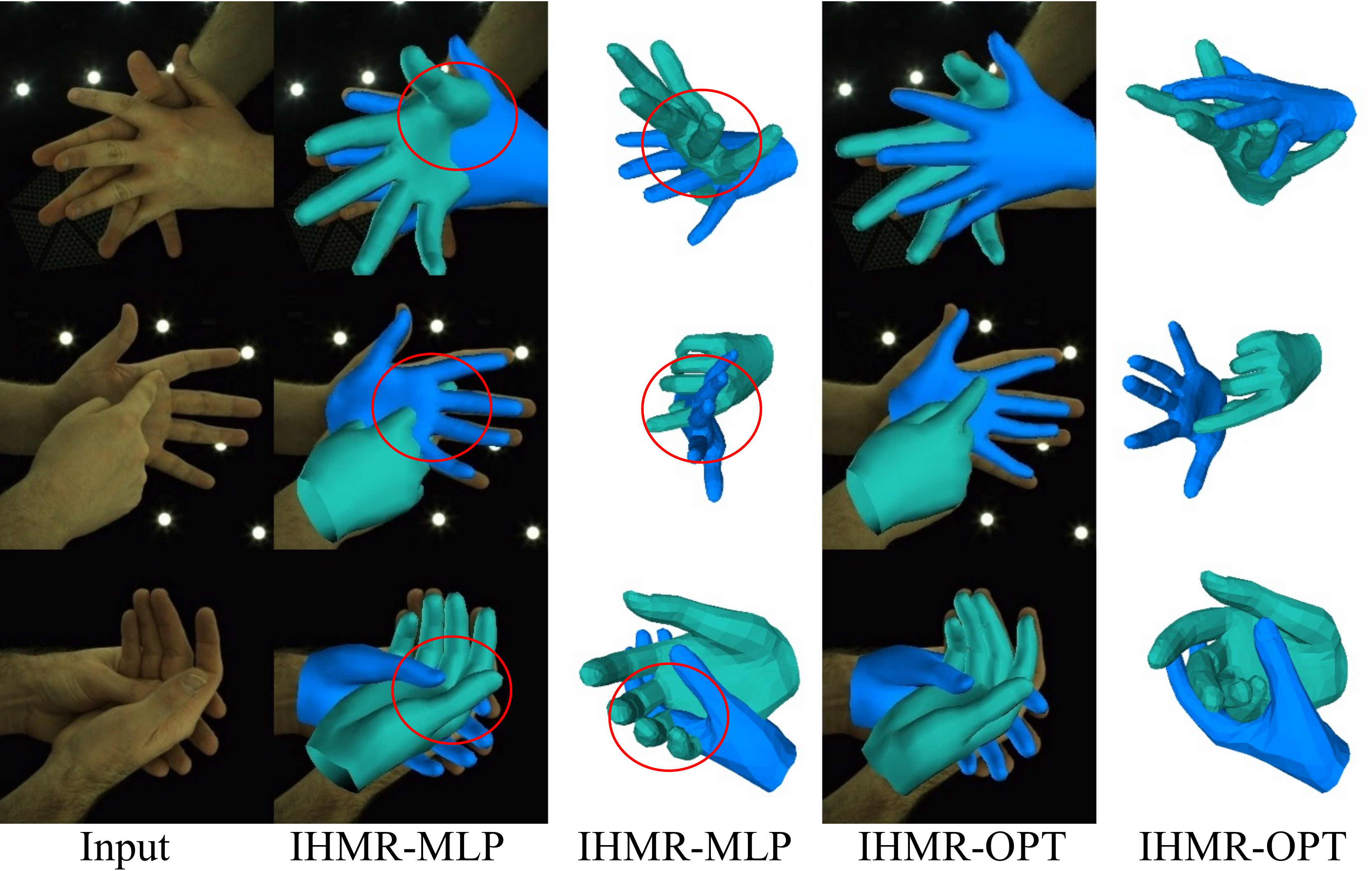}
	\end{center}
	\caption{\small \textbf{Qualitative Comparison between \methodname-MLP and \methodname-OPT.} We qualitatively compare between \methodname-MLP and \methodname-OPT. Collided regions are marked with red circles.}
	\label{fig:compare_mlp_opt}
\end{figure*}

\begin{figure*}[t]
	\begin{center}
		\includegraphics[width=0.95\linewidth]{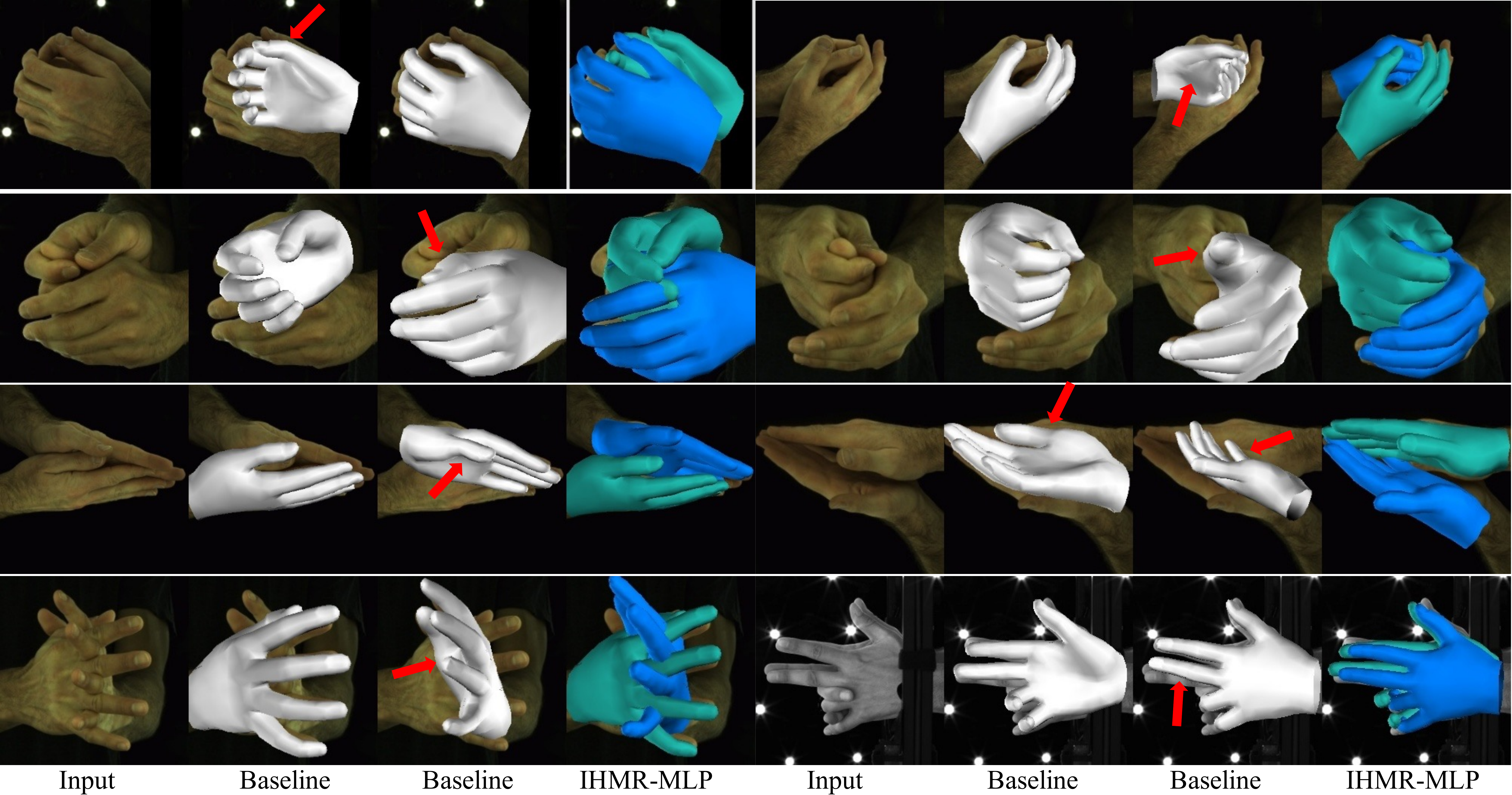}
	\end{center}
	\caption{\small \textbf{Compare with SOTA single-hand method.} In this figure, we qualitatively compare with single-hand baseline,~\ie Bouk~\cite{boukhayma20193d}. From left to right are input images, predicted right hands from the baseline, predicted left hands from the baseline and predicted interacting hands from~\methodname-MLP. Imprecise hand poses or finger poses generated by the single-hand method are marked with red arrows.}  
	\label{fig:qualitative_single_hand}
\end{figure*}

\begin{figure}[t]
	\begin{center}
		\includegraphics[width=0.85\linewidth]{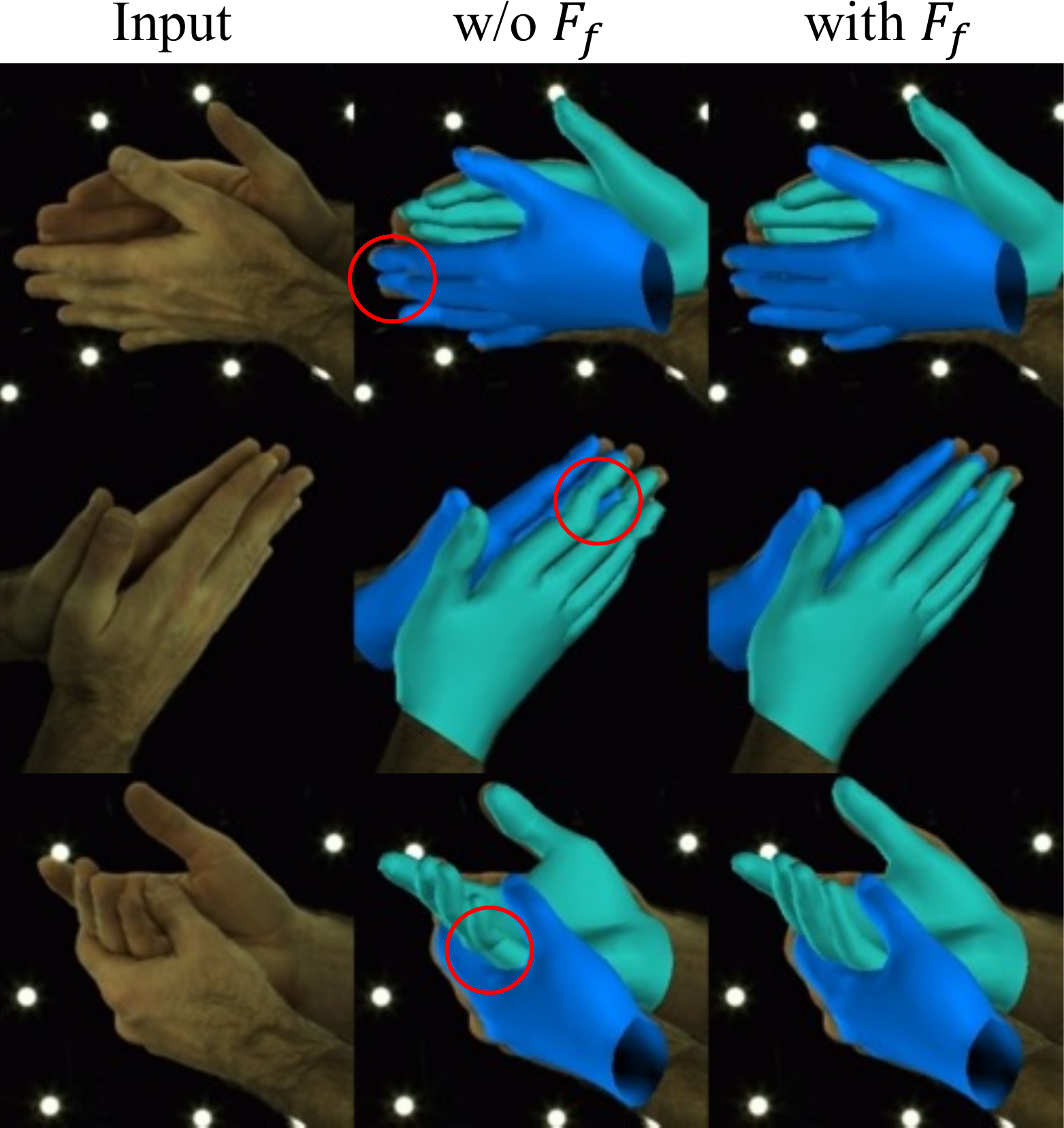}
	\end{center}
	\vskip -0.5cm
	\caption{\small \textbf{Role of $\mF_f$.} In this figure, we demonstrate the role of $\mF_f$ plays in optimization. From left to right are input images, optimized results without using $\mF_f$, optimized results with using $\mF_f$. Twisted fingers generated by optimization without using finger regularization $\mF_f$ are marked with red circles.}
	\label{fig:finger_reg}
\end{figure}

\begin{figure}[t]
	\begin{center}
		\includegraphics[width=0.9\linewidth]{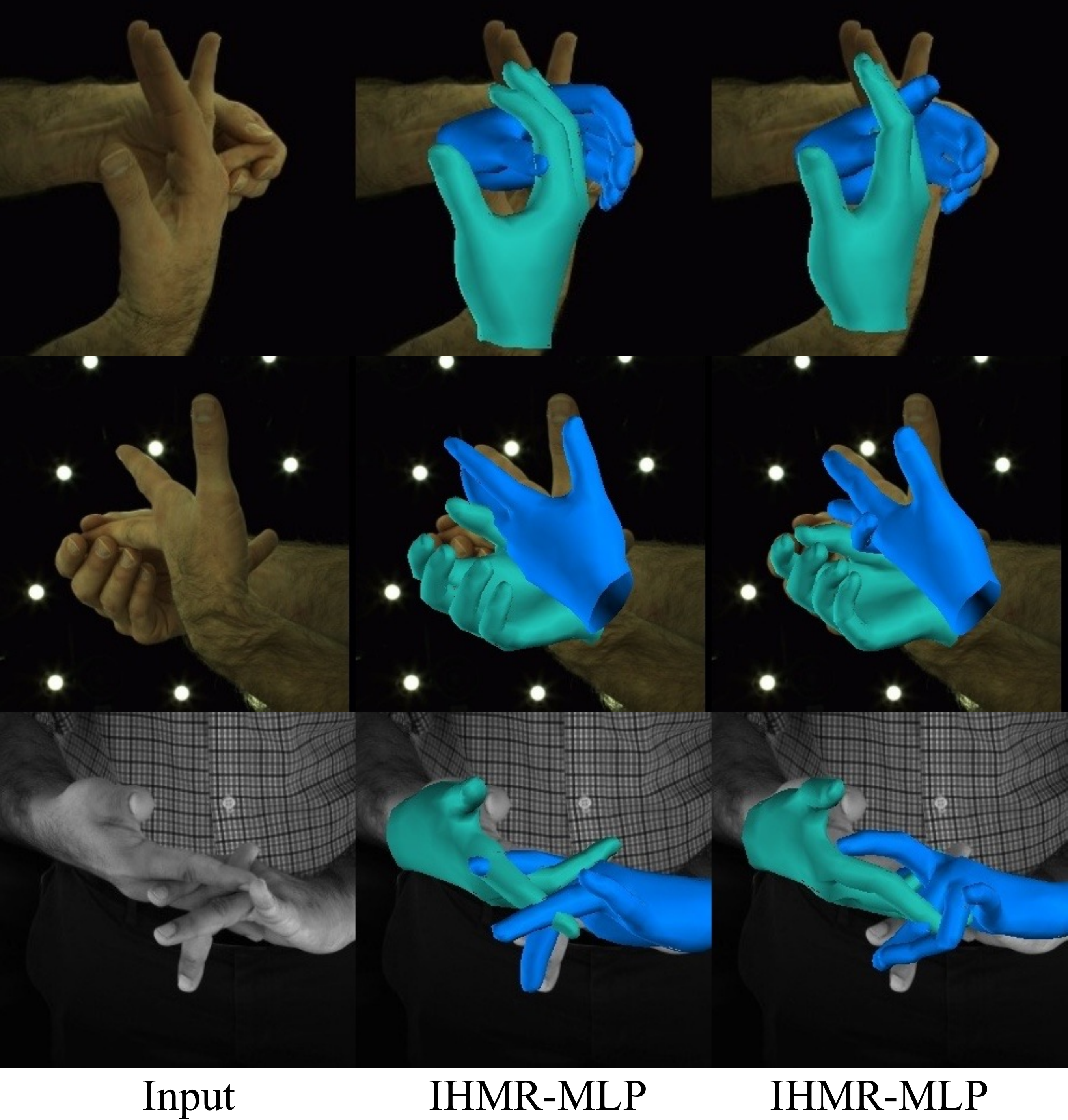}
	\end{center}
	\vskip -0.5cm
	\caption{\small \textbf{Typical Failure Cases.} Typical failure cases of \methodname are caused by challenging finger poses and collusions.}
	\label{fig:failure}
\end{figure}

\section{More Qualitative Results}
In this section, we show more qualitative results, including qualitative comparison between \methodname-MLP and \methodname-OPT, comparison between \methodname-MLP between single hand baseline, and quanlitative results demonstrating the effectiveness of finger regularization $\mF_f$, and typical failure cases.

\subsection{Comparison between Optimization and MLP}
We show quanlitative comparison between \methodname-OPT and \methodname-MLP in Fig.~\ref{fig:compare_mlp_opt}.
It is revealed that the optimization-based implementation can produce better 3D reconstructed hands with more precise joint estimation and fewer collisions.

\subsection{Compare with Single-hand Methods}
In this subsection, we qualitatively compare \methodname with the single-hand baseline,~\ie Bouk~\etal~\cite{boukhayma20193d}. The results are shown in Fig.~\ref{fig:qualitative_single_hand}.
It is revealed that the proposed {\methodname} can generate more precise interacting 3D hands than single-hand methods, which can only treat interacting hands as isolated single hands.

\subsection{Influence of Finger Regularization}
In this subsection, we show the influence of applying finger regularization $\mF_f$ defined in Equ.(12) of the main paper. Several examples are listed in Fig.~\ref{fig:finger_reg}. It is revealed that without applying $\mF_f$, optimization methods tend to generate twisted fingers as marked by red circles in Fig.~\ref{fig:finger_reg}.

\subsection{Failure Cases.}
We show several typical failure cases in Fig.~\ref{fig:failure}.
As the results show, typical failure cases are caused by challenging poses and occlusions.

\clearpage


%% file: tables/opt_implement.tex

\begin{table}[t] \centering 
	\caption{\small \textbf{Weights and step size used in ~\methodname-OPT.} $\gamma$ is the step size in optimizing the parameters. Objectives including $\mF_{col}$ \etc are defined in Equ.(10) of the main paper.}
	\setlength\tabcolsep{2.5 pt}
	\resizebox{\columnwidth}{!}{
		\begin{tabular}{c|ccccccc}
			\hline 
			Param $\hbrho$ $\downarrow$ & $\gamma$ & $\mF_{col}$  & $\mF_{2D}$ & $\mF_{3D}$ & $\mF_{\tau}$ & $\mF_{reg}$ & $\mF_{f}$ \\
			\hline
            $\hbtau$ & $1\mathrm{e}{-4}$ & $1\mathrm{e}{-1}$ & $1\mathrm{e}{+1}$ & $1\mathrm{e}{+3}$ & $1\mathrm{e}{+3}$ & $1\mathrm{e}{-1}$ & $0.0$ \\
            $\hbphi$ & $1\mathrm{e}{-2}$ & $1\mathrm{e}{-0}$ & $1\mathrm{e}{+1}$ & $1\mathrm{e}{+3}$ & $1\mathrm{e}{+3}$ & $1\mathrm{e}{-1}$ & $0.0$ \\
            $\hbtheta$ & $1\mathrm{e}{-2}$ & $1\mathrm{e}{-0}$ & $1\mathrm{e}{+1}$ & $1\mathrm{e}{+3}$ & $1\mathrm{e}{+3}$ & $1\mathrm{e}{-1}$ & $1\mathrm{e}{+5}$ \\
            $\hbbeta$ & $1\mathrm{e}{-2}$ & $1\mathrm{e}{-0}$ & $1\mathrm{e}{+1}$ & $1\mathrm{e}{+3}$ & $1\mathrm{e}{+3}$ & $1\mathrm{e}{-1}$ & $0.0$ \\
			\hline
		\end{tabular}	
	}
	\label{tab:opt_implement}
\end{table}

%% file: tables/compare_sota_mpvpe.tex

\begin{table*}[t]
	\begin{center}
		\caption{\small \tb{Comparison with SOTA methods on InterHand2.6M using Vertex-Based Method.}. We additionally evaluate on subsets of InterHand2.6M~\cite{moon2020interhand} using vertex-based metrics MPVPE and I-MPVPE. AVE-P and MAX-P are adopted to estimate the collision status of the generated interacting hands.}
		\vskip -0.2cm
		\setlength\tabcolsep{3 pt}
		\resizebox{2.1\columnwidth}{!}{
			\begin{tabular}{l|c|ccc|ccc}
				\hline
				Dataset \& Metrics $\rightarrow$   & IH26M  & \multicolumn{3}{c|}{IH26M-Inter} & \multicolumn{3}{c}{IH26M-Inter-Close}  \\
				\hline
				Methods $\downarrow$ 			   & MPJPE / MPVPE $\downarrow$ & MPJPE / MPVPE $\downarrow$ & I-MPJPE / I-MPVPE $\downarrow$ & AVE-P / MAX-P $\downarrow$  & MPJPE / MPVPE $\downarrow$ & I-MPJPE / I-MPVPE $\downarrow$  & AVE-P / MAX-P $\downarrow$ \\
                \hline
                Bouk~\etal~\cite{boukhayma20193d}  & 21.96 / 18.88 & 22.55 / 18.86 & - &  - & 21.20 / 18.81 & - & - \\
			    Pose2Mesh~\cite{choi2020pose2mesh} & 21.76 / 18.61 & 22.73 / 18.91 & - &  - & 20.82 / 18.37 & - & - \\
                BiHand~\cite{yang2020bihand}       & 19.90 / 17.17 & 21.18 / 17.32 & - &  - & 19.80 / 17.47 & -  & - \\
				\hline
				\methodname-Baseline 		       & 21.67 / 17.54 & 22.60 / 17.62 & 24.38 / 25.08 & 0.45 / 9.98 & 21.24 / 17.56 & 18.25 / 18.60 & 0.84 / 14.40  \\
				\methodname-MLP      			   & 22.79 / 18.16 & 23.37 / 17.82 & 21.26 / 21.85 & 0.33 / 8.03 & 21.55 / 17.58 & 16.60 / 16.82 & 0.68 / 12.67  \\
				\methodname-OPT 				   & 19.04 / 16.94 & 24.09 / 18.82 & 16.82 / 17.23 & \tb{0.13} / \tb{3.75} & 19.04 / 16.94 & 15.40 / 15.33 & \tb{0.33} / \tb{7.30} \\
				\hline
				\methodname-Baseline$^{*}$         & 17.05 / 17.18 & 17.54 / 16.71 & 14.44 / \tb{12.45} & 0.29 / 5.76 & 16.91 / 16.53 & 13.16 / 11.47 & 0.69 / 10.72 \\
				\methodname-MLP$^{*}$ 			   & 15.68 / \tb{14.57} & \tb{16.45} / \tb{14.69} & 13.64 / 11.81 & 0.26 / 5.34 & 15.76 / \tb{14.56} & 12.46 / \tb{10.77} & 0.61 / 10.07 \\
				\methodname-OPT$^{*}$			   & \tb{15.47} / 17.17 & 16.52 / 16.44 & \tb{13.49} / 13.01 & 0.23 / 4.48 & \tb{15.32} / 14.73 & \tb{11.90} / 11.23 & 0.34 / 7.48  \\
				\hline
			\end{tabular}
		}
		\vspace {-0.2cm}
		\label{tab:compare_sota_mpvpe}
	\end{center}
\end{table*}

%% file: tables/ablation_objective.tex


\begin{table}[t] \centering 
	\caption{\small \textbf{Role of different optimization objectives.} We study the role of three objectives used in the optimization-based factorized refinement, namely 3D objectives $\mF_{3D}$, 2D objectives $\mF_{2D}$ and objective on hand translation $\mF_{\tau}$.}
	\vskip -0.2cm
	\resizebox{0.8\columnwidth}{!}{
		\begin{tabular}{ccc|cc}
			\hline
			$\mF_{2D}$ & $\mF_{3D}$ & $\mF_{\tau}$ & I-MPJPE $\downarrow$ & AVE-P $\downarrow$ \\
			\hline
			- & - & - & 20.29 & 0.70 \\
			\hline
			\cmark & & & 18.99 & 0.14  \\
			& \cmark & &  17.07 &  0.22 \\
			& & \cmark & 36.76 & \textbf{0.01} \\
			\hline 
			\cmark & \cmark & & \textbf{16.06} & 0.27 \\
			\cmark & & \cmark & 17.89 & 0.12 \\
			& \cmark & \cmark & 16.99 & 0.21 \\
			\cmark & \cmark & \cmark & 16.94 & 0.20 \\
			\hline
		\end{tabular}	
	}
	\vspace{-0.0cm}
	\label{tab:ablation_objective}
\end{table}

%% file: tables/ablation_joint.tex

\begin{table}[t] \centering 
	\caption{\small \textbf{Influence of the quality of pseudo-ground-truth 3D joints.} We add noise to ground-truth 3D joints and use these joints to serve as the pseudo 3D joints, namely $\dbj_{3D}$. For still joints, we still use the original pseduo-ground-truth 2D joints, $\dbj_{2D}$.}
	\vskip -0.3cm
	\resizebox{\columnwidth}{!}{
		\begin{tabular}{c|cccc}
			\hline 
			\mrow{STD of Noise (mm) $\downarrow$}  & \multicolumn{2}{c}{I-MPJPE $\downarrow$} & \multicolumn{2}{c}{AVE-P $\downarrow$} \\
			& OPT & MLP & OPT & MLP \\
			\hline 
			$0$    	 					& 15.13 & 17.91 & 0.183 & 0.525 \\
			$10$						& 15.27 & 18.16 & 0.152 & 0.515 \\
			$20$						& 15.56	& 18.37	& 0.170	& 0.503	\\
			$30$						& 16.10 & 18.96	& 0.178 & 0.527 \\
			$40$						& 16.72	& 19.14	& 0.181 & 0.495 \\
			\hline 
		\end{tabular}	
	}
	\vspace{-0.3cm}
	\label{tab:ablation_joint}
\end{table}